\documentclass[conference]{IEEEtran}
\IEEEoverridecommandlockouts
\usepackage{url}
\usepackage{cite}

\usepackage{algorithm}
\usepackage{amsmath,amssymb,amsfonts}
\usepackage{algorithmic}
\usepackage{graphicx}
\usepackage{textcomp}
\usepackage{xcolor}
\usepackage{mathabx}
\usepackage{bm}
\newcommand{\bs}[1] {\bm{#1}}
\usepackage{multirow}

\ifCLASSOPTIONcompsoc
 \usepackage[caption=false,font=normalsize,labelfont=sf,textfont=sf]{subfig}
\else
 \usepackage[caption=false,font=footnotesize]{subfig}
\fi

\def\BibTeX{{\rm B\kern-.05em{\sc i\kern-.025em b}\kern-.08em
    T\kern-.1667em\lower.7ex\hbox{E}\kern-.125emX}}
\begin{document}

\title{Neural Plasticity Networks}

\author{\IEEEauthorblockN{Yang Li}
\IEEEauthorblockA{\textit{Department of Computer Science} \\
\textit{Georgia State University}\\
Atlanta, USA \\
yli93@student.gsu.edu}
\and
\IEEEauthorblockN{Shihao Ji}
\IEEEauthorblockA{\textit{Department of Computer Science} \\
\textit{Georgia State University}\\
Atlanta, USA \\
sji@gsu.edu}}

\maketitle

\begin{abstract}
    Neural plasticity is an important functionality of human brain, in which number of neurons and synapses can shrink or expand in response to stimuli throughout the span of life. We model this dynamic learning process as an $L_0$-norm regularized binary optimization problem, in which each unit of a neural network (e.g., weight, neuron or channel, etc.) is attached with a stochastic binary gate, whose parameters determine the level of activity of a unit in the network. At the beginning, only a small portion of binary gates (therefore the corresponding neurons) are activated, while the remaining neurons are in a hibernation mode. As the learning proceeds, some neurons might be activated or deactivated if doing so can be justified by the cost-benefit tradeoff measured by the $L_0$-norm regularized objective. As the training gets mature, the probability of transition between activation and deactivation will diminish until a final hardening stage. We demonstrate that all of these learning dynamics can be modulated by a single parameter $k$ seamlessly. Our neural plasticity network (NPN) can prune or expand a network depending on the initial capacity of network provided by the user; it also unifies dropout (when $k=0$), traditional training of DNNs (when $k=\infty$) and interpolates between these two. To the best of our knowledge, this is the first learning framework that unifies network sparsification and network expansion in an end-to-end training pipeline. Extensive experiments on synthetic dataset and multiple image classification benchmarks demonstrate the superior performance of NPN. We show that both network sparsification and network expansion can yield compact models of similar architectures, while retaining competitive accuracies of the original networks.\footnote{Our source code is available at \url{https://github.com/leo-yangli/npns}.}
  \end{abstract}
  
  
  %
  \IEEEpeerreviewmaketitle

  \section{Introduction} 
  Deep Neural Networks (DNNs) have achieved great success in a broad range of applications in image recognition~\cite{imagenet09}, natural language processing~\cite{bert18}, and game playing~\cite{alphago16}. Along with this success is a paradigm shift from feature engineering to architecture design. Latest DNN architectures, such as ResNet~\cite{resnet16}, DenseNet~\cite{densenet17} and Wide-ResNet~\cite{zagoruyko2016wide}, incorporate hundreds of millions of parameters to achieve state-of-the-art predictive performance. However, the expanding number of parameters not only increases the risk of overfitting, but also leads to high computational costs. A practical solution to this problem is network sparsification, by which weights, neurons or channels can be pruned or sparsified significantly with minor accuracy losses~\cite{han2015learning,han2015deep}, and sometimes sparsified networks can even achieve higher accuracies due to the regularization effects of the network sparsification algorithms~\cite{NekMolAsh17,Louizos2017}. Driven by the widespread applications of DNNs in resource-limited embedded systems, there has been an increasing interest in sparsifying networks recently~\cite{han2015learning,han2015deep,WenWuWan16,li2016pruning,louizos2017bayesian,molchanov2017variational,NekMolAsh17,Louizos2017}. 
  
  A less explored alternative is network expansion, by which weights, neurons or channels can be gradually added to grow a small network to improve its predictive accuracy. This paradigm is in the opposite to network sparsification, but might be more desirable because (1) we don't need to set an upper-bound on the network capacity (e.g., number of weights, neurons or channels, etc.) to start with, and the network can shrink or expand as needed for a given task; (2) it's computationally more efficient to train a small network and expand it to a larger one as redundant neurons are less likely to emerge during the whole training process; and (3) network expansion is more biologically plausible than network sparsification according to our current understanding to human brain development~\cite{brain}.
  
  In this paper, we propose Neural Plasticity Networks (NPNs) that unify network sparsification and network expansion in an end-to-end training pipeline, in which number of neurons and synapses can shrink or expand as needed to solve a given learning task. We model this dynamic learning process as an $L_0$-norm regularized binary optimization problem, in which each unit of a neural network (e.g., weight, neuron or channel, etc.) is attached with a stochastic binary gate, whose parameters determine the level of activity of a unit in the whole network. The activation or deactivation of a unit is completely data-driven as long as doing so can be justified by the cost-benefit tradeoff measured by the $L_0$-norm regularized objective. As a result, given a network architecture either large or small, our NPN can automatically perform sparsification or expansion without human intervention until a suitable network capacity is reached. 
  
  NPN is built on top of our previous $L_0$-ARM algorithm~\cite{l0-arm}. However, the original $L_0$-ARM algorithm only explores network sparsification, in which it demonstrates state-of-the-art performance at pruning networks, while here we extend this framework to network expansion. On the algorithmic side, we further investigate the Augment-Reinforce-Merge (ARM)~\cite{Yin2019}, a recently proposed unbiased gradient estimator for binary latent variable models. We show that due to the flexibility of ARM, many smooth or non-smooth parametric functions, such as scaled sigmoid or hard sigmoid, can be used to parameterize the $L_0$-norm regularized binary optimization problem and the unbiasness of the ARM estimator is retained, while a closly related hard concrete estimator~\cite{louizos2017bayesian} has to rely on the hard sigmoid function for binary optimization. It is this difference that entails NPN the capability of shrinking or expanding network capacity as needed for a given task. We also introduce a learning stage scheduler for NPN and demonstrate that many training stages of network sparsification and expansion, such as pre-training, sparsification/expansion and fine-tuning, can be modulated by a single parameter $k$ seamlessly; along the way, we also give a new interpretation of dropout~\cite{dropout}. Extensive experiments on synthetic dataset and multiple public datasets demonstrate the superior performance of NPNs for network sparsification and network expansion with fully connected layers, convolutional layers and skip connections. Our experiments show that both network sparsification and network expansion can converge to similar network capacities with similar accuracies even though they are initialized with networks of different sizes. To the best of our knowledge, this is the first learning framework that unifies network sparsification and network expansion in an end-to-end training pipeline modulated by a single parameter.
  
  The remainder of the paper is organized as follows. In Sec.~\ref{sec:formulation} we describe the $L_0$-norm regularized empirical risk minimization for NPN and its solver $L_0$-ARM~\cite{l0-arm} for network sparsification. A new learning stage scheduler for NPN is introduced in Sec.~\ref{sec:scheduler}. We then extend NPN to network expansion in Sec.~\ref{sec:expansion}, followed by the related work in Sec.~\ref{sec:related}. Experimental results are presented in Sec.~\ref{sec:exp}. Conclusions and future work are discussed in Sec.~\ref{sec:conclusion}.
  
  \section{Neural Plasticity Networks: Formulation}\label{sec:formulation}
  Our Neural Plasticity Network (NPN) is built on the basic framework of $L_0$-ARM~\cite{l0-arm}, which was proposed in our previous work for network sparsification. We extend $L_0$-ARM to network expansion, and unify network sparsification and expansion in an end-to-end training pipeline. For the sake of clarity, we first introduce NPN in the context of network sparsification, and later extend it to network expansion. The formulation below largely follows that of $L_0$-ARM~\cite{l0-arm}.
  
  Given a training set $D = \left\{ \left( \bs { x } _ { i } , y_i \right) , i = 1,2 , \cdots , N \right\}$, where $\bs { x_i }$ denotes the input and $y_i$ denotes the target, a neural network is a function $h(\bs x; \bs { \theta })$ parametrized by $\bs { \theta }$ that fits to the training data $D$ with the goal of achieving good generalization to unseen test data. To optimize $\bs { \theta }$, typically a regularized empirical risk is minimized, which contains two terms -- a data loss over training data and a regularization loss over model parameters. Empirically, the regularization term can be weight decay or Lasso, i.e., the $L_2$ or $L_1$ norm of model parameters. 
  
  Intuitively, network sparsification or expansion is a model selection problem, in which a suitable model capacity is selected for a given learning task. In this problem, how to measure model complexity is a core issue. The Akaike Information Criterion (AIC)~\cite{AIC} and the Bayesian Information Criterion (BIC)~\cite{BIC}, well-known model selection criteria, measure model complexity by counting number of non-zero parameters. Since the $L_2$ or $L_1$ norm only imposes shrinkage on large values of $\bs{\theta}$, the resulting model parameters $\bs{\theta}$ are often manifested by smaller magnitudes but none of them are exact zero. Therefore, the $L_2$ or $L_1$ norm is not suitable for measuring model complexity. A more appealing alternative is the $L_0$ norm of model parameters as it measures \textit{explicitly} the number of non-zero parameters, which is the \textit{exact} model complexity measured by AIC and BIC. With the $L_0$ regularization, the empirical risk objective can be written as
      \begin{equation}\label{eq:risk}
          \mathcal { R } ( \bs { \theta } ) = \frac { 1 } { N } \sum _ { i = 1 } ^ { N } \mathcal { L } \left( h ( \bs { x } _ { i } ; \bs { \theta } ) , y_i \right) + \lambda \| \bs { \theta } \| _ { 0 }
      \end{equation} 
  where $\mathcal { L } (\cdot)$ denotes the data loss over training data $D$, such as the cross-entropy loss for classification or the mean squared error (MSE) for regression, and $\| \bs { \theta } \| _ { 0 } $ denotes the $L_0$-norm over model parameters, i.e., the number of non-zero weights, and $\lambda$ is a regularization hyperparameter that balances between data loss and model complexity. For network sparsification, minimizing of Eq.~\ref{eq:risk} will drive the redundant or insignificant weights to be exact zero and thus pruned away. For network expansion, adding additional neurons will increase model complexity (the second term) but potentially can reduce data loss (the first term) and therefore the total loss. Thus, we will use Eq.~\ref{eq:risk} as our guiding principle for sparsifying or expanding a network. 
  
  To represent a sparsified network, we attach a binary random variable $z$ to each element of model parameters $\bs{\theta}$. Therefore, we can reparameterize the model parameters $\bs{\theta}$ as an element-wise product of non-zero parameters $\tilde{\bs{\theta}}$ and binary random variables $\bs{z}$:
      \begin{equation}\label{eq:theta}    
          \bs { \theta }  = \bs { \tilde { \theta } }  \odot \bs { z },
      \end{equation} 
  where $ \bs z  \in \{ 0,1 \} ^ { | \bs { \theta } | } $, and $\odot$ denotes the element-wise product. As a result, Eq.~\ref{eq:risk} can be rewritten as:
  \begin{align}\label{eq:R}
          \mathcal { R } ( \tilde { \bs { \theta } }, \bs{z} ) = \frac { 1 } { N } \sum _ { i = 1 } ^ { N } \mathcal { L } \left( h \bigl( \bs { x } _ { i } ; \tilde { \bs { \theta } } \odot \bs { z } \bigr) , y_i \right) + \lambda \sum _ { j = 1 } ^ { | \bs { \tilde { \theta } } | } \bs { 1 }_{\left[  z_j \neq 0 \right]},
  \end{align}
  where $\bs { 1 }_{[c]}$ is an indicator function that is $1$ if the condition $c$ is satisfied, and $0$ otherwise. Note that both the first term and the second term of Eq.~\ref{eq:R} are not differentiable w.r.t. $\bs { z }$. Therefore, further approximations need to be considered.
  
  Fortunately, we can approximate Eq.~\ref{eq:R} through an inequality from stochastic variational optimization~\cite{SVO18}. Specifically, given any function $ \mathcal { F } (\bs{z})  $ and any distribution $q(\bs{z})$, the following inequality holds
  \begin{equation}\label{eq:Ele}
          \min_{\bs{z}} \mathcal { F } (\bs{z}) \leq \mathbb{E}_{\bs{z}\sim q(\bs{z})} [\mathcal { F }(\bs{z})],
      \end{equation} 
  i.e., the minimum of a function is upper bounded by the expectation of the function. With this result, we can derive an upper bound of Eq.~\ref{eq:R} as follows.
  
  Since $z_j, \forall j\in\{1,\cdots,|\bs{\theta}|\}$ is a binary random variable, we assume $z_j$ is subject to a Bernoulli distribution with parameter $\pi_j\in[0, 1]$, i.e. $z_j\sim \mathrm { Ber } (z;\pi_j)$. Thus, we can upper bound $\min_{\bs{z}} \mathcal { R } ( \tilde { \bs { \theta } }, \bs{z} )$ by the expectation
  \begin{align}\label{eq:rhat}
  &\mathcal {\hat{R}} ( \tilde { \bs { \theta } } , \bs { \pi } ) 
    =\mathbb { E }_ { \bs { z } \sim  \mathrm{ Ber  } ( \bs { z } ; \bs { \pi } ) } \mathcal { R } ( \tilde { \bs { \theta } } , \bs{z}) \\ 
          & = \mathbb { E } _ { \bs { z } \sim  \mathrm{ Ber } ( \bs { z } ; \bs { \pi } ) } \left[ \frac { 1 } { N } \sum _ { i = 1 } ^ { N } \mathcal { L } \left( h ( \bs { x } _ { i } ; \tilde { \bs { \theta } } \odot \bs { z } ) , y_i \right)  \right] + \lambda \sum _ { j = 1 } ^ { | \tilde{\bs\theta} | } \pi _ { j }. \nonumber
  \end{align}
  As we can see, now the second term is differentiable w.r.t. the new model parameters $\bs {\pi} $, while the first term is still problematic since the expectation over a large number of binary random variables $\bs {z}\in\{0,1\}^{|\theta|}$ is intractable, so is its gradient. 
  
  
  To minimize Eq.~\ref{eq:rhat}, $L_0$-ARM utitlizes the Augment-Reinforce-Merge (ARM)~\cite{Yin2019}, an unbiased gradient estimator, to this stochastic binary optimization problem. Specifically, 
  
  \textbf{Theorem 1} (ARM)~\cite{Yin2019}. For a vector of $V$ binary random variables $\bs { z }=\left(z_{1}, \cdots, z_{V}\right)$, the gradient of 
      \begin{equation}
          \mathcal { E } ( \bs { \phi } ) = \mathbb { E } _ { \bs { z } \sim \prod _ { v = 1 } ^ { V } \operatorname { Ber } \left( z _ { v } ; g \left( \phi _ { v } \right) \right) } [ f ( \bs { z } ) ]
      \end{equation} 
  w.r.t. $\bs { \phi} = (\phi_1, \cdots, \phi_V)$, the logits of the Bernoulli distribution parameters, can be expressed as 
  \begin{align}\label{eq:arm}
    \nabla _ { \phi } \mathcal { E } ( \phi ) 
    = &\mathbb { E } _ { \bs { u } \sim \prod _ { v = 1 } ^ { V }\!\!  \operatorname { Uniform } \left( u _ { v } ; 0,1 \right) }\Big[ \big( f ( \bs { 1 } _ {[ \bs { u } > g ( - \phi ) ] }) -  \nonumber \\
    &f ( \bs { 1 } _ { [ \bs { u } < g ( \bs { \phi } ) ] } ) \big) ( \bs { u } - 1 / 2 )\Big ],
  \end{align}
  where $\bs { 1 } _ { [ \bs { u } > g ( - \bs { \phi } ) ] } : = \bs { 1 } _ { \left[ u _ { 1 } > g \left( - \phi _ { 1 } \right) \right] } , \cdots , \bs { 1 } _ { \left[ u _ { V } > g \left( - \phi _ { V } \right) \right] }$ and $g(\phi)=\sigma(\phi)=1/(1+\exp(-\phi))$ is the sigmoid function.   
  
  Parameterizing $\pi_j\in[0, 1]$ as $g {(\phi _ { j })}$, we can rewrite Eq.~\ref{eq:rhat} as
  \begin{align}\label{eq:L0-obj}
              \mathcal {\hat{R} } ( \tilde { \bs { \theta } } , \bs { \phi } ) &=  \mathbb { E }_ { \bs { z } \sim \mathrm{ Ber } ( \bs { z } ; g (\bs { \phi } ) ) } \left[f(\bs { z })\right] + \lambda \sum _ { j = 1 } ^ { | \tilde{\theta} | } g {(\bs { \phi }  _ { j })}  \\
              &=  \mathbb { E }_ { \bs { u } \sim \mathrm { Uniform } ( \bs { u } ; 0, 1) } \left[f(\bs { \bs { 1 } _ { [ \bs { u } < g ( \bs { \phi } ) ] } })\right] + \lambda \sum _ { j = 1 } ^ { | \tilde{\theta} | } g {(\bs { \phi }  _ { j })},\nonumber
  \end{align}
  where $f(\bs{z}) = \frac { 1 } { N } \sum _ { i = 1 } ^ { N } \mathcal { L } \left( h ( \bs { x } _ { i } ; \tilde { \bs { \theta } } \odot \bs { \bs{z} }) , y_i \right) $.
  From Theorem 1, we can evaluate the gradient of Eq.~\ref{eq:L0-obj} w.r.t. $\bs{\phi}$ by
  \begin{align}\label{eq:L0-ARM}
    &\!\! \nabla_{ \bs { \phi } } \mathcal { \hat{R} } (\tilde { \bs { \theta } }, \bs{\phi} ) = \mathbb { E } _ { \bs { u } \sim \mathrm { Uniform } \left( \bs{u}; 0,1 \right)} \Big[ \big( f ( \bs { 1 } _ { [ \bs { u } > g ( - \bs{\phi} ) ] } ) - 
              \nonumber\\
              & \quad f ( \bs { 1 } _ { [ \bs { u } < g ( \bs { \phi } ) ] }) \big) ( \bs { u } - 1 / 2)\Big] + \lambda \sum _ { j = 1 } ^ { | \tilde{\theta} | } \nabla_{\phi_j}g {(\phi_{ j })},
  \end{align}
  which is an unbiased and low variance estimator as demonstrated in~\cite{Yin2019}.
  
  
  \subsection{Choice of $g(\phi)$}\label{sec:choice}
\begin{figure}[t]
      \centering
      \includegraphics[width=0.75\linewidth]{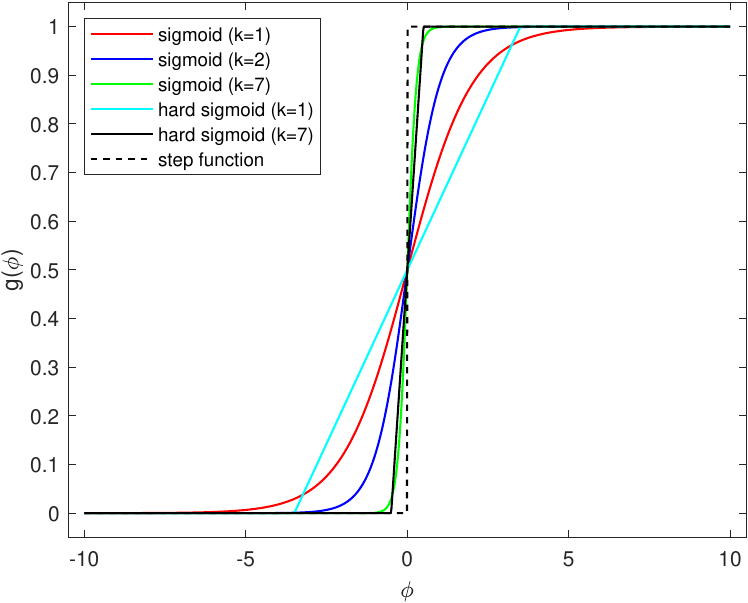}\vspace{-5pt}
      \caption{The plots of $g(\phi)$ with different $k$ for sigmoid and hard sigmoid functions. A large $k$ tends to be more effective at sparsifying networks. Best viewed in color.}\label{fig:sigmoid}
    \end{figure}

  Theorem 1 of ARM defines $g(\phi)=\sigma(\phi)$, where $\sigma(\cdot)$ is the sigmoid function. For the purpose of network sparsification and expansion, we find that this parametric function isn't very effective due to its fixed rate of transition between values 0 and 1. Thanks to the flexibility of ARM, we have a large freedom to design this parametric function $g(\phi)$. Apparently, it's straightforward to generalize Theorem 1 for any parametric functions (smooth or non-smooth) as long as $g: \mathcal{R}\to[0, 1]$ and $g(-\phi)=1-g(\phi)$\footnote{The second condition is not necessary. But for simplicity, we will impose this condition to select parametric function $g(\phi)$ that is antithetic. Designing $g(\phi)$ without this constraint could be a potential area that is worthy of further investigation.}. Example parametric functions that work well in our experiments are the scaled sigmoid function 
  \begin{align}\label{eq:sigmoid}
    g_{\sigma_k}(\phi) = \sigma(k\phi)=\frac{1}{1+\exp(-k\phi)},
  \end{align}
  and the centered-scaled hard sigmoid
  \begin{align}\label{eq:hard_sigmoid}
    g_{\bar{\sigma}_k}(\phi) = \min(1, \max(0, \frac{k}{7}\phi+0.5)),
  \end{align}
  where $7$ is introduced such that $g_{\bar{\sigma}_1}(\phi)\!\approx\! g_{\sigma_1}(\phi)\!=\!\sigma(\phi)$. See Fig.~\ref{fig:sigmoid} for some example plots of $g_{\sigma_k}(\phi)$ and $g_{\bar{\sigma}_k}(\phi)$ with different $k$s. Empirically, we find that $k\!=\!7$ works well for network sparsification, and $k\!=\!0.5$ for network expansion. More on this will be discussed when we present results. 
  
  One important difference between the hard concrete estimator from Louizos et al.~\cite{Louizos2017} and $L_0$-ARM is that the hard concrete estimator has to rely on the hard sigmoid gate to zero out some parameters during training (a.k.a. conditional computation~\cite{BenLeoCou18}), while $L_0$-ARM achieves conditional computation naturally by sampling from the Bernoulli distribution, parameterized by $g(\phi)$, where $g(\phi)$ can be any parametric function (smooth or non-smooth) as shown in Fig.~\ref{fig:sigmoid}. The consequence of using the hard sigmoid gate is that once a unit is pruned, the corresponding gradient will be always zero due to the exact 0 gradient at the left tail of the hard sigmoid gate (see Fig.~\ref{fig:sigmoid}) and therefore it can never be reactivated in the future. To mitigate this issue of the hard concrete estimator, $L_0$-ARM can utilize the scaled sigmoid gate~(\ref{eq:sigmoid}), which has non-zero gradient everywhere $(-\infty,\infty)$, and therefore a unit can be activated or deactivated freely and thus be plastic.

  \section{Learning Stage Scheduler}\label{sec:scheduler}
  As far as we know, all network sparsification algorithms either operate in a three-stage of pre-training, sparsification, and fine-tuning~\cite{han2015deep,han2015learning,WenWuWan16} or only have one sparsification stage from scratch~\cite{Louizos2017}. It has been shown that the three-stage sparsification leads to better predictive accuracies than the one-stage alternatives~\cite{LeeKimYoo18}. To support this three-stage learning process, previous methods~\cite{han2015deep,han2015learning,WenWuWan16} however manage this tedious process manually. Thanks to the flexibility of NPN, we can modulate these learning stages by simply adjusting $k$ of the $g_k(\phi)$ function at different training stages. Along the way, we also discover a new interpretation of dropout~\cite{dropout}.
  
  \subsection{Dropout as $k=0$}
  When $k = 0$, it is readily to verify that $g_{\sigma_k}(\phi) = g_{\bar{\sigma}_k}(\phi) = 0.5$, and the objective function~(\ref{eq:L0-obj}) is degenerated to 
  \begin{align}\label{eq:L0-obj-dropout}
    \mathcal {\hat{R} } ( \tilde { \bs { \theta } } , \bs { \phi } ) =  \mathbb { E }_ { \bs { u } \sim \mathrm { Uniform } ( \bs { u } ; 0, 1) } \left[f(\bs { \bs { 1 } _ { [ \bs { u } < 0.5 ] } })\right] + \lambda  | \tilde{\theta} |/2,
  \end{align}
  which is in fact the standard dropout with a dropout probability of 0.5~\cite{dropout}. Note that the value of 0.5 is due to the artifact of the antithetic constraint on the parametric function $g(\phi)$. As we discussed in Sec.~\ref{sec:choice}, this constraint isn't necessary and we have freedom of designing $g(0)=c$ with $c\in[0,1]$, which corresponds to any dropout probability of the standard dropout. From this point of view, dropout is just a special case of NPN when $k=0$, and this is a new interpretation of dropout.
  
  \subsection{Pre-training as $k=\infty$ at the beginning of NPN training}
  At the beginning of NPN training, we initialize all $\phi$'s to some positive values, e.g., $\phi>0.1$. If we set $k = \infty$, then $g_{\sigma_{\infty}}(\phi) = g_{\bar{\sigma}_{\infty}}(\phi) = 1$ and the objective function~(\ref{eq:L0-obj}) becomes 
  \begin{align}\label{eq:L0-obj-FM}
    \mathcal {\hat{R} } ( \tilde { \bs { \theta } } , \bs { \phi } ) =  \mathbb { E }_ { \bs { u } \sim \mathrm { Uniform } ( \bs { u } ; 0, 1) } \left[f(\bs { \bs { 1 } _ { [ \bs { u } < 1 ] } })\right] + \lambda  | \tilde{\theta} | ,
  \end{align}
  which corresponds to the standard training of DNNs with all neurons activated. Moreover, the gradient w.r.t. $\phi$ is degenerated to
  \begin{align}\label{eq:L0-ARM-pretrain}
    \nabla_{ \bs { \phi } } \mathcal { \hat{R} } (\tilde { \bs { \theta } }, \bs{\phi} ) &= \mathbb { E } _ { \bs { u } \sim \mathrm { Uniform } \left( \bs{u}; 0,1 \right)} \Big[ \big( f ( \bs { 1 } _ { [ \bs { u } > 0 ] } ) 
    \\ &- f ( \bs { 1 } _ { [ \bs { u } < 1 ] }) \big) ( \bs { u } - 1 / 2)\Big] + \lambda \sum _ { j = 1 } ^ { | \tilde{\theta} | } \nabla_{\phi_j} 1 = 0, \nonumber
  \end{align}
  such that $\phi$ will not be updated during the training and the architecture is fixed. This corresponds to the pre-training of a network from scratch.
  
  \subsection{Fine-tuning as $k=\infty$ at the end of NPN training}
  At the end of NPN training, the histogram of $g(\phi)$ is typically split to two spikes of values around 0 and 1 as demonstrated in $L_0$-ARM~\cite{l0-arm}. If we set $k = \infty$, then the values of $g(\phi)$ will be exactly 0 or 1. In this case, the gradient w.r.t. $\phi$ is zero, the neurons with $g(\phi)=1$ are activated and the neurons with $g(\phi)=0$ are deactivated. This corresponds to the case of fine-tuning a fixed architecture without the $L_0$ regularization.
  
  \subsection{Modulating learning stages by $k$} \label{three-stages}
  As discussed above, we can now integrate the three-stage of pre-training, sparsification and fine-tuning into one end-to-end pipeline, modulated by a single parameter $k$. At the beginning of the training, we set $k=\infty$ to pre-train a network from scratch. Upon convergence, we can set $k$ to some small values (e.g., $k=7$) to enable the $L_0$ regularized optimization for network sparsification. After the convergence, we can set $k=\infty$ again to fine-tune the final learned architecture without the $L_0$ regularization. To the best of our knowledge, there is no other network sparsification algorithm that supports this three-stage training in a native end-to-end pipeline. As an analogy to the common learning rate scheduler, we call $k$ as a learning stage scheduler. We will demonstrate this when we present results.
  
  \section{Network Expansion}\label{sec:expansion}
  So far we have described NPN in the context of network sparsification. Thanks to the flexibility of $L_0$-ARM, it is straightforward to extend it to network expansion. Instead of starting from a large network for pruning, we can expand a small network by adding neurons during training, an analogy of growing a brain from small to large. Specifically, given the level of activity of a neuron is determined by its $\phi$, a new neuron can be added to the network with a large $\phi$ such that it will be activated in future training epochs. If this neuron is useful at reducing the $L_0$-regularized loss function~(\ref{eq:L0-obj}), its $\phi$ value will be increased such that it will be activated more often in the future; otherwise, it will be gradually deactivated and pruned away in the future. At each training iteration, we add a new neuron to a layer if (a) the validation loss is lower than its previous value when a neuron was added last time, and (b) all neurons in the layer are activated (i.e., no redundant neurons in the layer). The network expansion will terminate when the $L_0$-regularized loss plateaus or some added neurons are deactivated due to the $L_0$-norm regularization. This network expansion procedure is detailed in Algorithm~\ref{algo:expansion}.
  
  \begin{algorithm}
    \caption{Network Expansion}
    \begin{algorithmic}
    \REQUIRE number of iterations $T$, deep network $f$, number of layers $L$, $loss_{old} = +\infty$, $loss_{val} = 0$
    \FOR{$t = 1$ to $T$}
      \STATE Train $f$ on a mini-batch of training examples
      \STATE $loss_{val} =$ Validate $f$ on validation dataset
      \IF {$loss_{val} < loss_{old}$}
        \FOR {$l = 1$ to $L$ }
          \IF {all neurons in layer $l$ are activated} 
            \STATE Add a new neuron to layer $l$
          \ENDIF
        \ENDFOR
        
        \STATE $loss_{old} = loss_{val}$
      \ENDIF
    \ENDFOR
    \end{algorithmic}
    \label{algo:expansion}
  \end{algorithm}
  
  Thanks to the learning stage scheduler discussed in Sec.~\ref{sec:scheduler}, we can simulate network expansion easily by manipulating $k$. Specifically, we can initialize a very large network to represent an upper-bound on network capacity. To start with a small network, we randomly select a small number of neurons and initialize the corresponding $\phi$s to large positive values and set all the remaining $\phi$'s to large negative values, such that only a small portion of the neurons will be activated while the remaining neurons are in a hibernation mode. Since they will not be activated, these hibernating neurons consume no computation resources. To pretrain the initial small network, we can train NPN with $k=\infty$. Upon convergence, we can randomly activate a few hibernating neurons (under the conditions discussed above) and switch $k$ to a small value (e.g., $k=0.5$). Along with the original neurons, we can optimize the expanded network by reducing the $L_0$-regularized loss. In such a way, we can readily simulate network expansion. Upon the network expansion terminates, we can set $k=\infty$ to fine-tune the final network architecture. In the experiments, we will resort to this approach to simulate network expansion.

  \section{Related Work}\label{sec:related}
  Our NPN has a built-in support to network sparsification, network expansion, and can automatically determine an appropriate network capacity for a given learning task. In this section, we review related works in these areas.
   
  \paragraph{Network Sparsification}
  Driven by the widespread applications of DNNs in resource-limited embedded systems, recently there has been an increasing interest in network sparsification~\cite{han2015learning,han2015deep,WenWuWan16,li2016pruning,louizos2017bayesian,molchanov2017variational,NekMolAsh17,Louizos2017,l0-arm}. One of the earliest sparsification methods is to prune the redundant weights based on the magnitudes~\cite{lecun1990optimal}, which is proved to be effective in modern CNNs~\cite{han2015learning}. Although weight sparsification is able to compress networks, it can barely improve computational efficiency due to unstructured sparsity. Therefore, magnitude-based group sparsity is proposed~\cite{WenWuWan16,li2016pruning}, which can prune networks while reducing computation cost significantly. These works are mainly based on the $L_2$ or $L_1$ regularization to penalize the magnitude of weights. A more appealing approach is based on the $L_0$ regularization~\cite{Louizos2017,l0-arm} as this corresponds to the well-known model selection criteria such as AIC~\cite{AIC} and BIC~\cite{BIC}. Our NPN is built on the basic framework of $L_0$-ARM~\cite{l0-arm} and extend it for network expansion. In addition, as far as we know almost all the network sparsification algorithms~\cite{han2015learning,han2015deep,WenWuWan16,li2016pruning} usually proceed in three stages manually: pretrain a full network, prune the redundant weights or filters, and fine-tune the pruned network. In contrast, our NPN can support this three-stage training by simply adjusting the learning stage scheduler $k$ at different stages in an end-to-end fashion. 
  
  \paragraph{Neural Architecture Search}
  Another closely related area is neural architecture search~\cite{ZopLe17,ZopVasShl18,ReaAggHua19} that searches for an optimal network architecture for a given learning task. It attempts to determine number of layers, types of layers, layer configurations, different activation functions, etc.
  Given the extremely large search space, typically reinforcement learning algorithms are utilized for efficient implementations. Our NPN can be categorized as a subset of neural architecture search in the sense that we start with a fixed architecture and aim to determine an optimal capacity (e.g., number of weights, neurons or channels) of a network.
  
  \paragraph{Dynamic Network Expansion}
  Compared to network sparsification, network expansion is a relatively less explored area. There are few existing works that can dynamically increase the capacity of network during training. For example, DNC~\cite{DNC89} sequentially adds neurons one at a time to the hidden layers of network until the desired approximation accuracy is achieved. \cite{ZhoSohLee12} proposes to train a denoising autoencoder (DAE) by adding in new neurons and later merging them with other neurons to prevent redundancy. For convolutional networks, \cite{WanRamHeb17} proposes to widen or deepen a pretrained network for better knowledge transfer. Recently, a boosting-style method named AdaNet~\cite{adanet17} is used to adaptively grow the structure while learning the weights. However, all these approaches either only add neurons or add/remove neurons manually. In contrast, our NPN can add or remove (deactivate) neurons during training as needed without human intervention, and is an end-to-end unified framework for network sparsification and expansion. 
  
  \section{Experimental Results}\label{sec:exp}
  We evaluate the performance of NPNs on multiple public datasets with different network architectures for network sparsification and network expansion. Specifically, we illustrate how NPN evolves on a synthetic ``moons" dataset~\cite{miyato2018virtual} with a 2-hidden-layer MLP. We also demonstrate LeNet5-Caffe\footnote{\url{https://github.com/BVLC/caffe/tree/master/examples/mnist}} on the MNIST dataset~\cite{mnist}, and ResNet56~\cite{resnet16} on the CIFAR10 and CIFAR100 datasets~\cite{krizhevsky2009learning}. Similar to $L_0$-ARM~\cite{l0-arm} and $L_0$-HC~\cite{louizos2017bayesian}, to achieve computational efficiency, only neuron-level (instead of weight-level) sparsification/expansion is considered, i.e., all weights of a neuron or filter are either pruned from or added to a network altogether. For the comparison to the state-of-the-art network sparsification algorithms~\cite{Louizos2017,molchanov2017variational,louizos2017bayesian}, we refer the readers to $L_0$-ARM~\cite{l0-arm} for more details since NPN is an extension of $L_0$-ARM for network sparsification and expansion.
  
  As discussed in Sec.~\ref{sec:formulation}, each neuron in an NPN is attached with a Bernoulli random variable parameterized by $g(\phi)$. Therefore, the level of activity of a neuron is determined by the value of $\phi$. To initialize an NPN, in our experiments we activate a neuron by setting $\phi=3/k$. Since $g_{\sigma_k}(\phi) = \sigma(3) \approx 0.95$, this means that the corresponding neuron has a 95\% probability of being activated. Similarly, we set $\phi = -3/k$ to deactivate a neuron with a 95\% probability.
  
  As discussed in Sec.~\ref{sec:scheduler}, all of our experiments are performed in three stages: (1) pre-training, (2) sparsification/expansion, and (3) fine-tuning, in an end-to-end training pipeline modulated by parameter $k$. In pre-training and fine-tuning stages, we set $k=5000$ (as a close approximation to $k=\infty$) to train an NPN with a fixed architecture. In sparsification/expansion stage, we set $k$ to a small value to allow NPNs to search for a suitable network capacity freely. 
  
  \begin{figure*}[t]
    \begin{center}
      \subfloat[Expansion: epoch 99]{\includegraphics[width=0.33\linewidth]{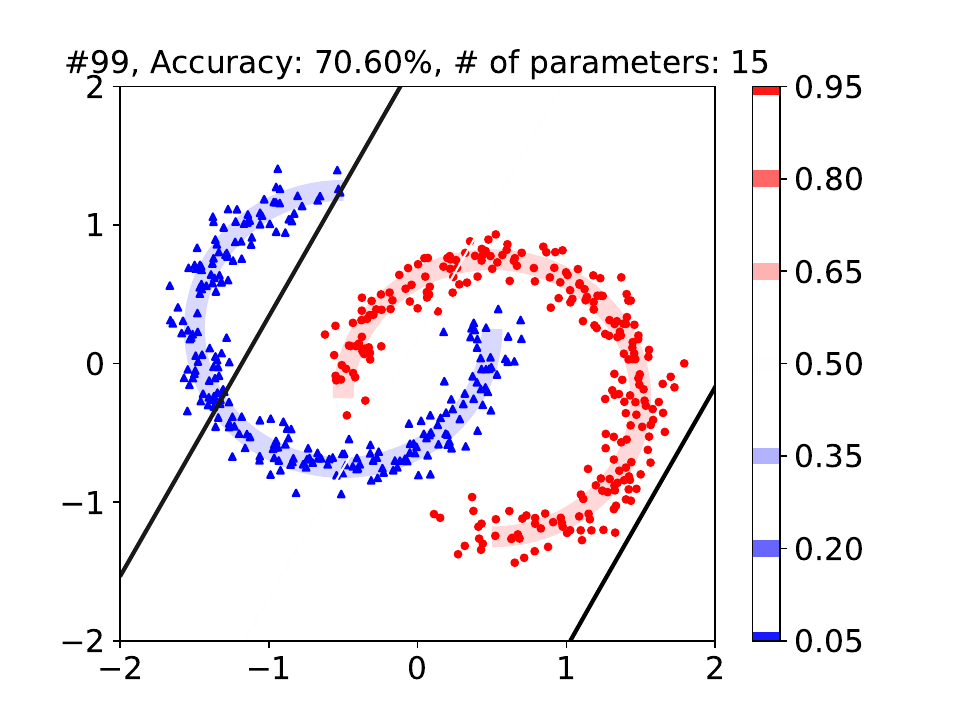}}\hfill
      \subfloat[Expansion: epoch 199]{\includegraphics[width=0.33\linewidth]{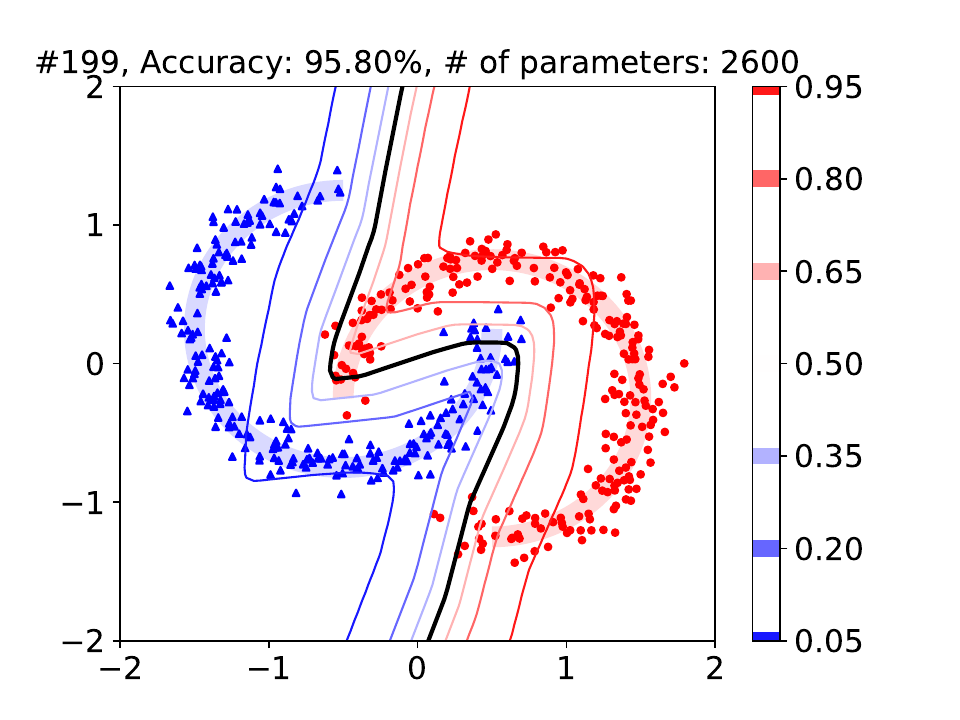}}\hfill
      \subfloat[Expansion: epoch 1999]{\includegraphics[width=0.33\linewidth]{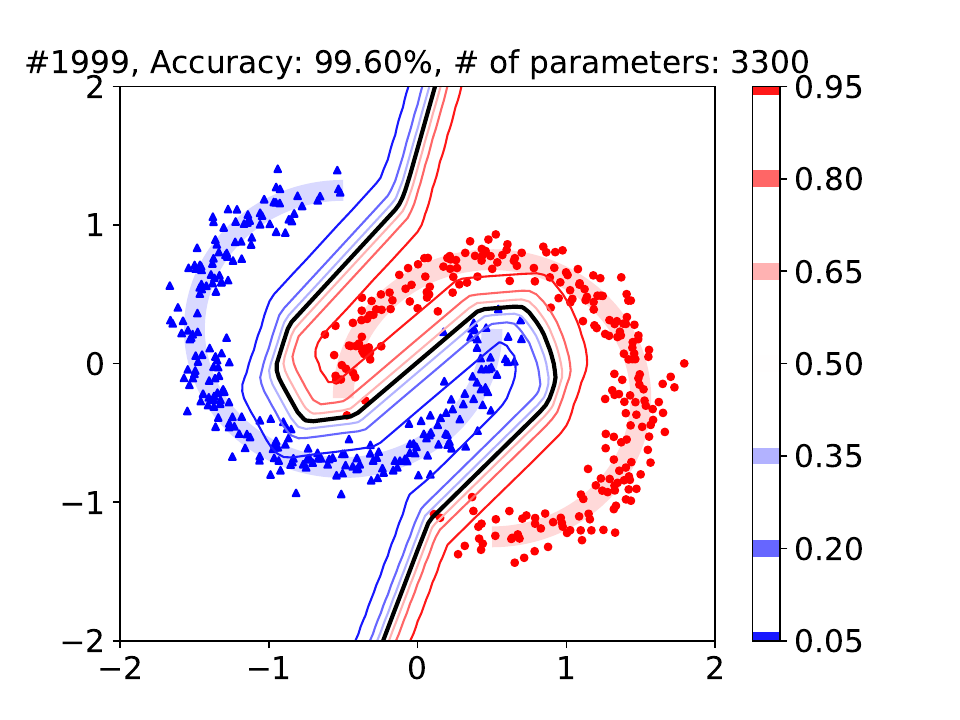}}\hfill
      \subfloat[Sparsification: epoch 499]{\includegraphics[width=0.33\linewidth]{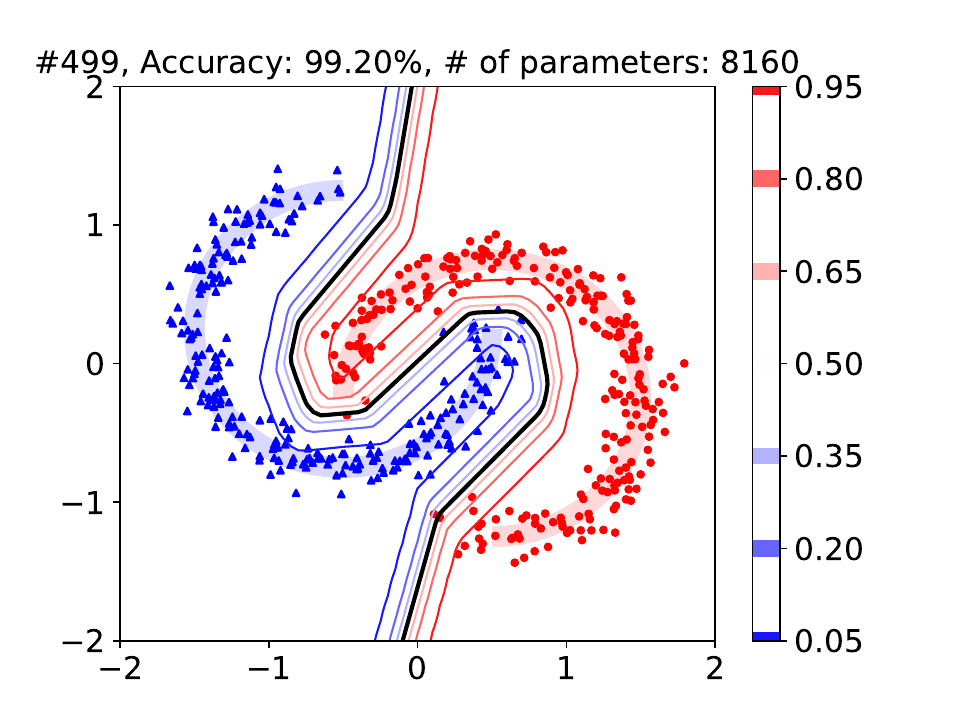}}\hfill
      \subfloat[Sparsification: epoch 999]{\includegraphics[width=0.33\linewidth]{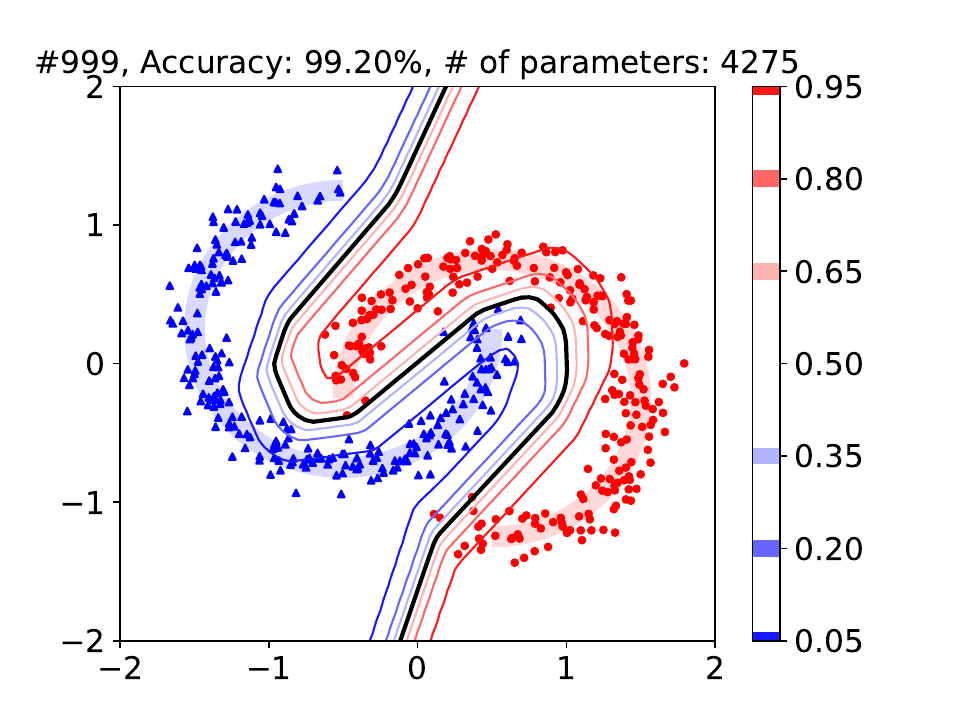}}\hfill
      \subfloat[Sparsification: epoch 1999]{\includegraphics[width=0.33\linewidth]{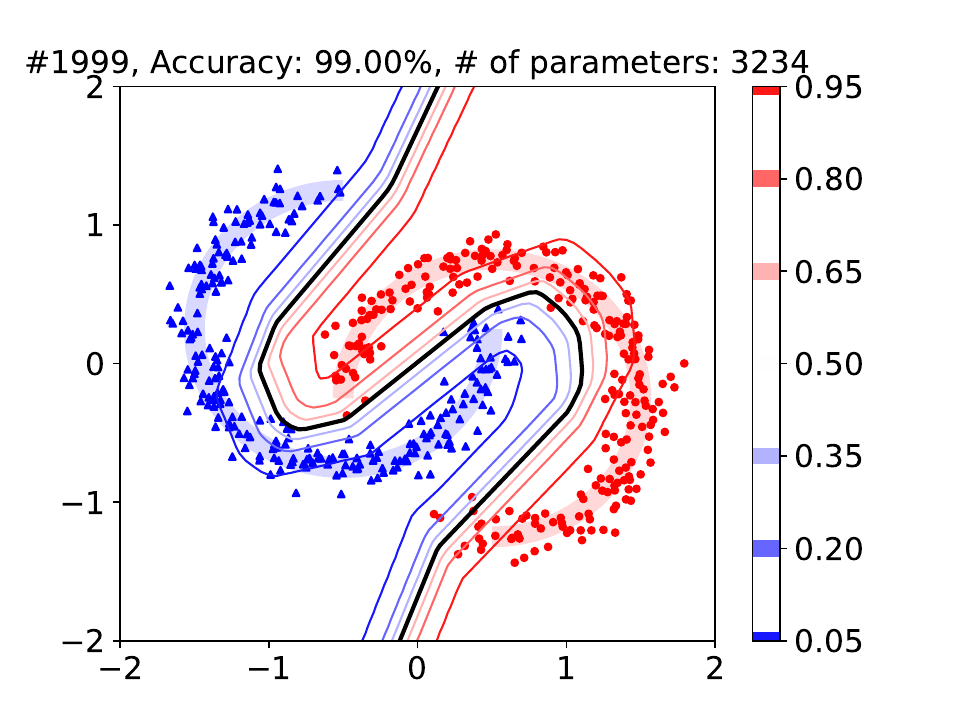}}\hfill
    \end{center}\vspace{-5pt}
    \caption{The evolution of the decision boundaries of NPNs for network expansion (a,b,c) and network sparsification (d,e,f). The
  videos can be found at https://npnsdemo.github.io.}\label{fig:toy}
  \end{figure*}
  
    The final architecture of a network is influenced significantly by two hyperparameters: (1) the regularization strength $\lambda$, and (2) $k$ of the $g_{\sigma_k}(.)$ function, which determine how aggressively to sparsify or expand a network. Typically, a positive $\lambda$ is used both for sparsification and expansion. However, in some expansion experiments, we notice that $\lambda=0$ is beneficial because $\lambda=0$ essentially encourages more neurons to be activated, which is important for network expansion to achieve competitive accuracies. For the hyperparameter $k$ used in stage 2, in all of our experiments we set $k = 7$ for sparsification and $k = 0.5$ for expansion. The reason that different $k$s are used is because for expansion we need to encourage the network to grow and a small $k$ is more amenable to keep neurons activated. The rest of hyperparameters of NPNs are determined via cross validation on the validation datasets.
  
  Unless specifically noted, we use the Adam optimizer~\cite{kingma2014adam} with an initial learning rate of 0.001. Our experiments are performed on NVIDIA Titan-Xp GPUs. Our source code is available at \url{https://github.com/leo-yangli/npns}.

  \subsection{Synthetic Dataset}
  
  To demonstrate that NPNs can adapt their capacities for a learning task, we visualize the learning process of NPNs on a synthetic ``moons" dataset~\cite{miyato2018virtual} for network sparsification and network expansion. The ``moons" dataset contains 1000 data points distributed in two moon-shaped clusters for binary classification. We randomly pick 500 data points for training and use the rest 500 data points for test. Two different MLP architectures are used for network sparsification and network expansion, respectively. For network sparsification, we train an MLP with 2 hidden layers of 100 and 80 neurons, respectively. The input layer has 2 neurons corresponding to the 2-dim coordinates of each data point, which is transformed to a 100-dim vector by a fixed matrix. The output layer has two neurons for binary classification. The overall architecture of the MLP is 2-100 (fixed)-80-2 with the first weight matrix fixed, and the overall number of trainable model parameters is 8,160 (excluding biases for clarity). The binary gates are attached to the outputs of the two hidden layers for sparsification. For the expansion experiment, we start an MLP with a very small architecture of 2-100 (fixed)-3-2. Initially, only three neurons at each hidden layer are activated, and therefore the total number of trainable model parameters is 15 (excluding biases for clarity). Apparently, the first MLP is overparameterized for this synthetic binary classification task, while the second MLP is too small and doesn't have enough capacity to solve the classification task with a high accuracy. 
  
  To visualize the learning process of NPNs,  in Fig.~\ref{fig:toy} we plot the decision boundaries and confidence contours of the NPN-sparsified MLP and NPN-expanded MLP on the test set of ``moons". We pick three snapshots from each experiment. The evolution of the decision boundaries of the NPN-expanded MLP is  shown in Fig.~\ref{fig:toy} (a, b, c). At the end of pre-training (a. epoch 99), the decision boundary is mostly linear and the capacity of the network is obviously not enough. During the stage 2 expansion (b. epoch 199), more neurons are added to the network and the decision boundary becomes more expressive as manifested by a piece-wise linear function, and at the same time the accuracy is significantly improved to about 96\%. At the end of stage 3 fine-tuning (c. epoch 1999), the accuracy reaches 99.2\% with more neurons being added. Similarly, Fig.~\ref{fig:toy} (d, e, f) demonstrates the evolution of the decision boundaries of NPN-sparsified MLP on the test set. The model achieves an accuracy of 99.2\% at the end of stage 1 pre-training (d. epoch 499). Then 47.6\% of weights are pruned without any accuracy loss during stage 2 sparsification (e. epoch 999). The model finally prunes 60.4\% of neurons at the end of stage 3 fine-tuning (f. epoch 1999). In this sparsification experiment, across different training stages, the shapes of decision boundaries are appropriately the same even though a large amount of neurons are pruned.
  
  Interestingly, the final architectures achieved by network expansion and network sparsification are very similar (3300 vs. 3234), so are their accuracies (99.6\% vs. 99.00\%) even though the initial network capacities are quite different (15 vs. 8160). This experiment demonstrates that given an initial network architecture either large or small, NPNs can adapt their capacities to solve a learning task with high accuracies. 
  
  \subsection{MNIST}
  In the second part of experiments, we run NPNs with LeNet5-Caffe on the MNIST dataset for network sparsification and expansion. LeNet5-Caffe consists of two convolutional layers of 20 and 50 neurons, respectively, interspersed with max pooling layers, followed by two fully-connected layers with 800 and 500 neurons. We start network sparsification from the full LeNet5 architecture (20-50-800-500, in short), while in the expansion experiment we start from a very small network with only 3 neurons in the first two convolutional layers, 48 and 3 neurons in the fully-connected layers (3-3-48-3, in short). We pre-train the NPNs for 100 epochs, followed by sparsification/expansion for 250 epochs and fine-tuning for 150 epochs. For both experiments, we use $\lambda=(10,0.5,0.1,10)/N$ where $N$ is the number of training images.
  
    \begin{table}[h]
    \caption{The network sparsification and expansion with LeNet5 on MNIST. The architecture and accuracy at the end of each stage are shown in the table.}
    \label{PLA_LENET}
    \centering
    \begin{tabular}{lccc}
      \hline
      & Stage  & Arch. (\# of Parameters) &  Accuracy (\%) \\
      \hline
      Baseline & - & 20-50-800-500 (4.23e5) & \textbf{99.40} \\
      \hline
      \multirow{3}*{Sparsification}  & stage 1 &  20-50-800-500 (4.23e5) & 99.36 \\
      ~ & stage 2 & 7-9-109-30 (5320)& 98.90 \\
      ~ & stage 3 & 7-9-109-30 (5320)& 98.91 \\
      \hline
      \multirow{3}*{Expansion}  & stage 1 & 3-3-48-3 (474) & 93.85 \\    
      ~ & stage 2 & 8-8-53-8 (2304)& 96.71\\
      ~ & stage 3 & 8-8-53-8 (\textbf{2304})& 98.31\\
      \hline
    \end{tabular}
  \end{table}
  
    \begin{figure*}[t]
    \begin{center}
      \subfloat[LeNet5: \# of parameters]{\includegraphics[width=0.4\linewidth]{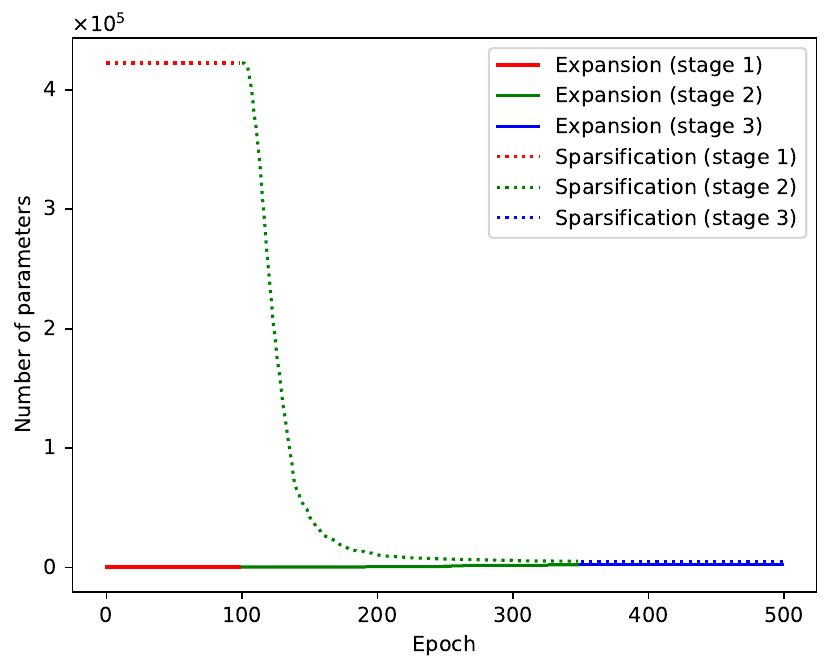}}\hspace{15pt}
      \subfloat[LetNet5: Test accuracy]{\includegraphics[width=0.412\linewidth]{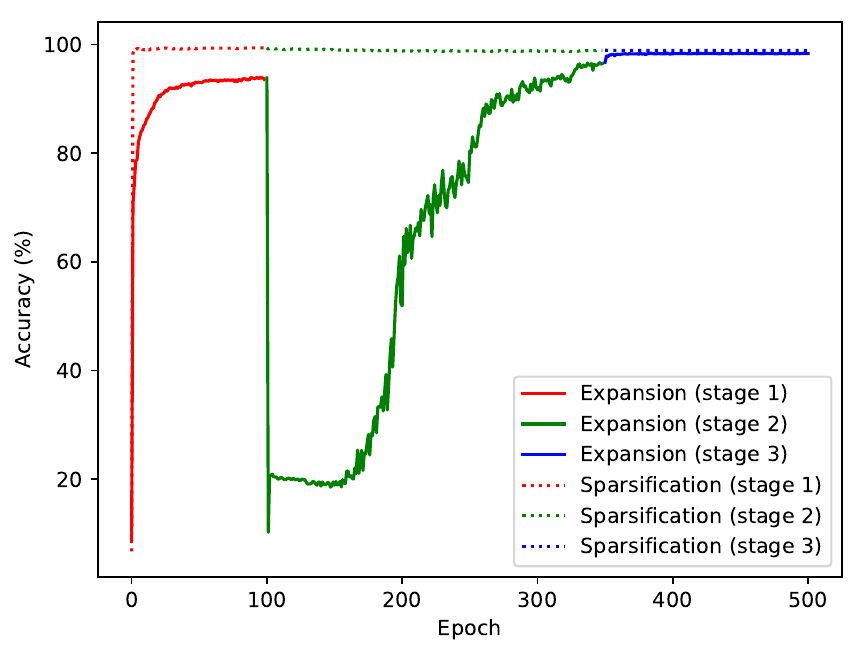}}
    \end{center}\vspace{-5pt}
    \caption{The evolution of network capacity and test accuracy as a function of epoch for NPN network sparsification and expansion with LetNet5 on MNIST.}\label{fig:mnist}
  \end{figure*}

  The results are shown in Table~\ref{PLA_LENET} and Fig.~\ref{fig:mnist}. For the sparsification experiment, the NPN achieves 99.36\% accuracy at the end of pre-training, and yields a sparse architecture with a minor accuracy drop at the end of sparsification. With the fine-tuning at stage 3, the accuracy reaches 98.91\% in the end. For the expansion experiment, the NPN achieves a low accuracy of 93.85\% after pre-training due to insufficient capacity of the initial network, then it expands to a larger network and improves the accuracy to 96.71\%. Finally, the accuracy reaches 98.31\% after fine-tuning. Fig.~\ref{fig:mnist} demonstrate the learning processes of NPNs on MNIST for network sparsification and network expansion. It is interesting to note that both network sparsification and expansion reach the similar network capacity with similar classification accuracies at the end of the training even though they are started from two significantly different network architectures. It's worth emphasizing that both sparsification and expansion reach similar accuracies to the baseline model, while over 99\% weights are pruned.
  
  \subsection{CIFAR-10/100}
  In the final part of experiments, we evaluate NPNs with ResNet56 on CIFAR-10 and CIFAR-100 for network sparsification and network expansion. Due to the existence of skip connections, we do not sparsify the last convolutional layer of each residual block to keep a valid addition operation. To train NPNs with this modern CNN architecture, two optimizers are used: (1) SGD with momentum for ResNet56 parameters with an initial learning rate of 0.1, and (2) Adam for the binary gate parameters $\bs{\phi}$ with an initial learning rate of 0.001. The batch size, weight decay and momentum of SGD are set to 128, 5e-4 and 0.9, respectively. The learning rate is multiplied by 0.1 every 60 epochs for SGD optimizer, while for Adam optimizer we multiplied learning rate by 0.1 at epoch 120 and 180. 
  
  As before, the networks are training in three stages by leveraging $k$. We pre-train the network for the first 20 epochs. In the expansion experiments, we pre-train a small network which only has 20\% of neurons, while in the sparsification experiments, we pre-triain the full architecture. The network is then trained in stage 2 for 180 epochs, and finally is fine-tuned in stage 3 for 20 epochs. For the sparsification experiments, we use $\lambda=1e-5$ for all layers, while for the expansion experiments we use $\lambda=0$ to encourage more neurons to be activated.

  \begin{table*}[ht]
  \small
  \centering
  \caption{The network sparsification and expansion with ResNet56 on CIFAR10 and CIFAR100. ``$\Delta$": `+' denotes accuracy gain; `-' denotes accuracy loss. ``Params. (P.R. \%)": prune ratio in parameters.}\vspace{-5pt}
  \begin{tabular}{l l c c c c}\hline
      Model & Method & Acc. (\%) & $\Delta_{Acc}$ & FLOPs (P.R. \%) & Params. (P.R. \%) \\\hline
      \multirow{7}{*}{CIFAR10} & 
      SFP~\cite{he2018soft} & 93.6$\rightarrow$93.4 & -0.2 & 59.4M (53.1) & - \\
       & AMC~\cite{he2018amc} & 92.8$\rightarrow$91.9 & \textcolor{red}{-0.9} & 62.5M (50.0) & - \\
       & FPGM~\cite{he2019filter} & 93.6$\rightarrow$93.5 & -0.1 & 59.4M (52.6) & - \\
       & TAS~\cite{dong2019network} & 94.5$\rightarrow$93.7 &\textcolor{red}{-0.8} & 59.5M (52.7) & - \\
       & HRank~\cite{lin2020hrank} & 93.3$\rightarrow$93.5 &\textbf{+0.2} & 88.7M (29.3) & 0.71M (16.8) \\     
      \cline{2-6}
       & NPN Sparsification & 93.2$\rightarrow$93.0  & -0.2 & 76.1M (40.1) & \textbf{0.33M (61.2)} \\
       & NPN Expansion  & 93.2$\rightarrow$92.7  & -0.5 & 100.5M (20.9) & 0.55M (35.3) \\
       \hline
       \multirow{5}{*}{CIFAR100} & SFP~\cite{he2018soft} & 71.4$\rightarrow$68.8 & \textcolor{red}{-2.6} & 59.4M (52.6) & -\\
      & FPGM~\cite{he2019filter} & 71.4$\rightarrow$69.7 & \textcolor{red}{-1.7} & 59.4M (52.6) & -\\
      & TAS~\cite{dong2019network} & 73.2$\rightarrow$72.3 & \textcolor{red}{-0.9} & 61.2M (51.3) & -\\
      \cline{2-6}
       & NPN Sparsification & 71.1$\rightarrow$70.9 & \textbf{-0.2} & 101.8M (19.8) & 0.61M (28.2) \\
       & NPN Expansion  & 71.1$\rightarrow$69.9 & -0.5 & 102.8M (19.1) & 0.60M (29.4)\\
       \hline
  \end{tabular}
  \label{tab:cifar}
\end{table*}
  
  We compare the performance of NPNs with the state-of-the-art pruning algorithms, including SFP~\cite{he2018soft}, AMC~\cite{he2018amc}, FPGM~\cite{he2019filter}, TAS~\cite{dong2019network} and HRank~\cite{lin2020hrank}, with the results shown in Table~\ref{tab:cifar}. Since the baseline accuracies in all the reference papers are different, we follow the common practice and compare the performances of all competing methods by their accuracy gains $\Delta_{Acc}$ and their pruning rates in terms of FLOPs and network parameters. It can be observed that NPN sparsification and NPN expansion achieve very competitive performances to the state-of-the-arts in terms of classification accuracies and prune rates. More interestingly, similar to the results on MNIST, both network sparsification and expansion reach the similar network capacities with similar classification accuracies at the end of the training even though they are started from two significantly different network architectures, demonstrating the plasticity of NPN for network sparsification and expansion.
  

  \section{Conclusion}\label{sec:conclusion}
  We propose neural plasticity networks (NPNs) for network sparsification and expansion by attaching each unit of a network with a stochastic binary gate, whose parameters are jointly optimized with original network parameters. The activation or deactivation of a unit is completely data-driven and determined by an $L_0$-regularized objective. Our NPN unifies dropout (when $k\!=\!0$), traditional training of DNNs (when $k\!\!=\!\!\infty$) and interpolate between these two. To the best of our knowledge, it is the first learning framework that unifies network sparsification and network expansion in an end-to-end training pipeline that supports pre-training, sparsification/expansion, and fine-tuning seamlessly. Along the way, we also give a new interpretation of dropout. Extensive experiments on multiple public datasets and multiple network architectures validate the effectiveness of NPNs for network sparsification and expansion in terms of model compactness and predictive accuracies. 
  
  As for future extensions, we plan to design better (possibly non-antithetic) parametric function $g(\phi)$ to improve the compactness of learned networks. We also plan to extend the framework to prune or expand network layers to further improve model compactness and accuracy altogether.
  
\section{Acknowledgment}  
The authors would like to thank the anonymous reviewers for their comments and suggestions, which helped improve the quality of this paper. The authors would also gratefully acknowledge the support of VMware Inc. for its university research fund to this research.

  \ifCLASSOPTIONcaptionsoff
    \newpage
  \fi

  
  
  \bibliographystyle{IEEEtran}
  \bibliography{npn_1915}

\end{document}